\pdfoutput=1                                            

\documentclass[utf8,twocolumn]{FrontiersinHarvard}

\usepackage{url,hyperref,microtype,subcaption}
\usepackage[onehalfspacing]{setspace}
\usepackage{multicol}
\usepackage{amsmath}
\usepackage{amssymb}
\usepackage{multirow}
\usepackage{comment}
\usepackage[utf8]{inputenc}

\def\keyFont{\fontsize{8}{11}\helveticabold }
\def\firstAuthorLast{Adra {et~al.}} 
\def\Authors{Mira Adra\,$^{1,*}$, Simone Melcarne\,$^{2}$, Nelida Mirabet-Herranz\,$^{2}$ and Jean-Luc Dugelay\,$^{2}$}
\def\Address{
$^{1}$GTD International, Ramonville-Saint-Agne, France \\
$^{2}$Dept. of Digital Security, EURECOM, Biot, France}
\def\corrAuthor{2 Rue Giotto, 31520 Ramonville-Saint-Agne}

\def\corrEmail{mira.adra@gtd.eu}

\begin{document}
\onecolumn
\firstpage{1}

\title[Event-based solutions for human-centered applications]{Event-based solutions for human-centered applications: A comprehensive review} 

\author[\firstAuthorLast ]{\Authors} 
\address{} 
\correspondance{} 

\extraAuth{}

\maketitle

\begin{abstract}


Event cameras, often referred to as dynamic vision sensors, are groundbreaking sensors capable of capturing changes in light intensity asynchronously, offering exceptional temporal resolution and energy efficiency. These attributes make them particularly suited for human-centered applications, as they capture both the most intricate details of facial expressions and the complex motion dynamics of the human body. Despite growing interest, research in human-centered applications of event cameras remains scattered, with no comprehensive overview encompassing both body and face tasks. This survey bridges that gap by being the first to unify these domains, presenting an extensive review of advancements, challenges, and opportunities. We also examine less-explored areas, including event compression techniques and simulation frameworks, which are essential for the broader adoption of event cameras. This survey is designed to serve as a foundational reference that helps both new and experienced researchers understand the current state of the field and identify promising directions for future work in human-centered event camera applications. A summary of this survey can be found at \url{https://github.com/nmirabeth/event_human}.

\tiny
 \keyFont{ \section{Keywords:} Neuromorphic sensors, Event-based vision, Human-centered applications, Event compression, Privacy preserving, Body analysis, Face analysis.} 
\end{abstract}

\section{Introduction}
Human-centered applications have long been at the forefront of computer vision, driven by the need to understand and analyze human activities in diverse contexts. Such applications span critical domains, including, among others, surveillance, where the objective is to monitor human behavior in real-time for public safety; biometric authentication, which can leverage unique individual features for secure identification; interactive systems that enable seamless human-computer interactions through gesture or expression recognition; and behavioral analysis, which provide insights into physical activity and social behaviors.


Although this research area encompasses a wide range of tasks, it can be broadly divided into two main directions of focus: body and face analysis. To this day, several traditional computer vision techniques have been developed to address both categories (\cite{viola2001rapid, zhang2016, Ma_2023, sun2019, yan2018spatialtemporalgraphconvolutional}). Such widely recognized methods  primarily rely on conventional RGB cameras and perform frame-based analysis, without considering or incorporating any other data types. However, these standard solutions are hindered by fundamental limitations. On the one hand, they are constrained by temporal resolution, meaning that they often fail to capture fast and subtle movements that, for example, distinguish micro-expressions or may happen in rapid gait changes. On the other hand, they are prone to motion blur in dynamic scenarios and struggle with challenging lighting conditions, such as high contrast or low illumination; finally, these approaches are resource-intensive as they require significant memory and processing power to handle high frame-rate video streams.

These challenges have recently fueled the interest in using neuromorphic (often referred to as dynamic vision sensors or event-based) cameras, which can offer a transformative solution. Event cameras are often called neuromorphic or bio-inspired sensors because they are modeled after the retina's sensory neurons, mimicking how photoreceptors respond to changes in light intensity rather than capturing static frames. Like the retina, event cameras operate asynchronously, with each pixel independently detecting brightness changes, much like sensory neurons and retinal ganglion cells in the visual system. This design is inspired by the transient pathways in biological vision, which specialize in detecting motion and dynamic changes in the environment. Just as the retina has cells dedicated to processing movement and contrast to help us perceive motion, event cameras replicate this functionality by focusing only on changes in brightness over time (\cite{steffen2019neuromorphic, posch2014retinomorphic}). Unlike the common belief that our eyes  "see" everything continuosly, the reality is that biological mechanisms such as microsaccades and the transient pathway ensure continuous perception by creating subtle changes in light input similar to the functioning of an event camera. Without these mechanisms, our vision would fade. Moreover, event cameras are also considered neuromorphic because they integrate seamlessly into neuromorphic computing frameworks, like spiking neural networks (SNNs), which mimic the way biological neurons transmit information as spikes, allowing for efficient, brain-like data processing. This design enables a series of unique characteristics that make them particularly well-suited for the aforementioned tasks.


This survey provides an overview of the current situation and the progress made in using event-based cameras specifically for human-centered applications, identifying key developments, existing challenges, and potential research directions to guide researchers at all levels.

\subsection{Related surveys}
Event-based cameras are being leveraged in a growing number of applications. Given their impact on both academia and industry, several surveys and reviews have been published in recent years, playing an important role in research as they summarize the state of the art, identify gaps, and propose directions for future investigation.
Some researchers have aimed at providing comprehensive overviews of the emerging field of event-based vision, describing in detail the physical sensor design and the technical specifications. In this direction, one of the earliest papers is represented by \cite{EtienneCummings1996}, a survey published more than two decades ago, which traces the history of neuromorphic sensor development. Since then, various surveys have been published over the years (\cite{kramer1998neuromorphic, indiveri2008neuromorphic, liu2010neuromorphic}), exploring topics such as hardware developments and the design of very-large-scale integration (VLSI) neuromorphic circuits for processing signals from event-based cameras. A more recent and exhaustive review is given by \cite{Gallego_2022}, which focused on event-based vision systems operating principles, underlying algorithms, and a wide range of applications addressed, mainly including robotics and perception. Similarly, \cite{chakravarthi2024recenteventcamerainnovations} investigated the latest innovations in event camera technology, examining models, datasets, and diverse applications across various domains, highlighting their impact on research and development. In addition, \cite{Cazzato2024EventBasedSurvey} proposed an application-driven survey, illustrating various outcomes across different application fields and exploring the issue of dataset availability.

In a different line, other studies have mainly focused on a specific topic, showing how event-based methods have evolved to tackle the challenges within that particular field. For example, in the targeted context of human-related analysis, \cite{Verschae2023EventGestureFacial} focused on event-driven gesture and facial expression recognition and compared different algorithms and benchmarks for the purpose of performance evaluation. \cite{becattini2024neuromorphicfaceanalysissurvey} discussed neuromorphic solutions for face analysis, which included detection, recognition, and emotion analysis and compared these with traditional approaches. Eye motion analysis with event cameras, and their potential for applications such as gaze estimation or blink detection, were explored by \cite{Iddrisu2024EventEyeMotion}. 

\subsection{Scope and value of our survey}

\begin{figure}[h!]
\begin{center}
\includegraphics[width=10cm]{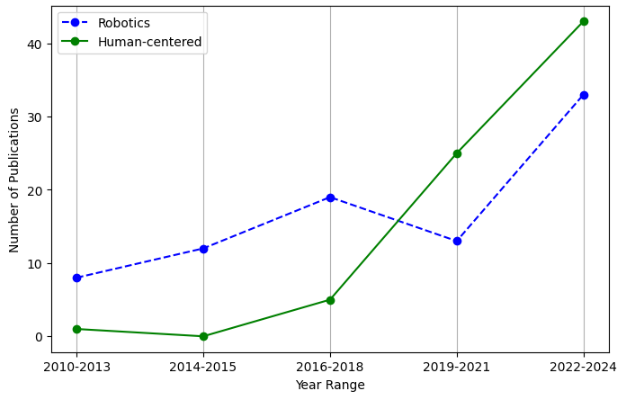}
\end{center}
\caption{ Evolution of research focus: Comparing the number of publications on robotics versus human-centered applications of event cameras. Results are based on searches conducted on Google Scholar.}\label{fig:1}
\end{figure}

The scope of this survey is specifically focused on human-centered applications of event cameras. These include applications addressing humans as a whole—such as gesture and action recognition, human tracking, and pose estimation—as well as applications focused on facial analysis, including face detection, emotion recognition, and face recognition. Unlike existing surveys, which often concentrate solely on face applications (\cite{becattini2024neuromorphicfaceanalysissurvey}) or narrowly on a subset of human actions (\cite{Verschae2023EventGestureFacial}), this survey aims to provide a comprehensive overview of all human-centered event-based applications. Our motivation stems from the evolution of event camera research trends, which have expanded beyond traditional robotics and high-speed tracking applications to demonstrate significant advantages in downstream human-centered tasks as highlighted in Figure \ref{fig:1}.
It is important to note that Figure \ref{fig:1} was created based on a focused methodology  based on papers with clear and explicit relevance to robotics and human-centered applications. The papers were selected from Google Scholar using specific keywords in their titles, such as 'robotics,' 'action recognition,' and 'face,' to ensure clarity and precision in the categorization.

What makes our survey particularly valuable is its uniqueness. To the best of our knowledge, this is the first survey to thoroughly target human-centered applications of event cameras, covering both body- and face-oriented use cases in a unified framework. Secondly, we want to emphasize the authors' contribution to the publications included in this survey, as our findings at Eurecom have contributed to advances in the field of neuromorphic computation across various human-centered applications for both body and face.

\subsection{Structure and coverage of our survey}
This survey is organized as follows: in Section \ref{Sec:Foundation}, we introduce the foundational concepts of event cameras, briefly explaining how they work, listing their main features and discussing the advantages and disadvantages of using this type of sensor in the specific context of human-centered applications; Section \ref{Sec:Event Data} explores the description of various strategies for representing event data, underlining the specificity that each of them has and for simulating event data starting from RGB videos using popular methods in the literature.
The following part presents a discussion on state-of-the-art datasets designed for experimental validation before highlighting some key techniques for event compression that enable efficient data handling; Section \ref{Sec:Applciations} provides an overview of the current state of the literature in the field of human-centered event-based applications, categorizing them into body- and face-related tasks; finally, in Section \ref{Sec:Conclusions}, we conclude with an analysis of current trends and offer future perspectives for the development and integration of event-based vision systems in real-world applications.

\section{Foundations of Event-Based Vision Systems} \label{Sec:Foundation}

Event cameras are bio-inspired vision sensors that represent a fundamental breakthrough compared to traditional frame-based imaging. Rather than capturing entire frames at regular intervals, these sensors operate in an \textit{asynchronous} manner, meaning that each pixel independently triggers an event only when a change in brightness exceeds a specific threshold. As a direct consequence, if no changes are detected, no data is generated, significantly reducing bandwidth usage. 

The properties of these cameras align with the dynamic and unpredictable characteristics of human activities, making them particularly suited for capturing fast and irregular actions, such as facial expressions or body movements. However, they also introduce unique challenges that require a deep rethinking of conventional vision methodologies.

In the following subsections, we provide the reader with a general overview of the functioning of event cameras, highlighting the advantages to be leveraged and the disadvantages to be managed.

\subsection{Operating Principle and Key Features}

A new \textit{event} $\textit{e}_k$ is triggered at a certain pixel $\textbf{u} = (u_k, v_k)$ whenever the change in brightness $\Delta L(\textbf{u}, t_k)$, calculated between the current time $t_k$ and the time of the last event $t_{k-1}$, exceeds a predefined contrast threshold $T>0$. Each event encodes 1) the pixel’s identity $\textbf{u} = (u_k, v_k)$ that indicate the location of the change, 2) a timestamp $\textit{t}_k$, capturing the precise time the event occurred, and 3) the polarity $\textit{p}_k \in \{-1, +1\}$, specifying whether the brightness increased or decreased; as a result, an event is represented as a tuple:




\begin{equation}
e_k = \{ \textbf{u}, t_k, p_k \}, \;
\label{eq:event}
\end{equation}
\begin{equation}
\textit{with }\Delta L(\textbf{u}, t_k) = L(\textbf{u}, t_k) - L(\textbf{u},t_{k-1}) \geq p_k T
\label{eq:threshold}
\end{equation}

In this sense, what a neuromorphic camera produces is nothing more than a spatio-temporally localized stream of events which can be formally described as: 

\begin{equation}
\mathcal{E} = \{ e_k \}_{k=1}^K \; \; \textit{where } k \in K,
\label{eq:stream of events}
\end{equation}

with  $K$ representing the total number of events occurring within the entire recording time interval.
This data-driven design makes the output rate dependent on scene dynamics,  since faster motion or more significant brightness variations lead to a higher event rate.
As reported in \cite{Gallego_2022}, several key features distinguish these sensors:

\textit{High Temporal Resolution}, since events are time-stamped with microsecond ($\mu s$) precision and allow the detection of rapid actions or subtle gestures, without motion blur; this ability at sensing vert fast motion is invaluable for tasks requiring fine-grained temporal analysis, such as gait analysis 
or blink detection. 


\textit{Low Latency}, as events are generated and transmitted as soon as brightness changes occur ensuring real-time responsiveness without the need to wait for a global frame exposure time; this is an essential requirement for applications like human tracking. For example, the DVS128 camera outputs events at a rate of 1 million events per second (Meps) while the Samsung DVS-Gen4 has a higher bandwidth of 1066 Meps.

\textit{High Dynamic Range (HDR)}, since event cameras can capture scenes with a vast range of lighting conditions, from dark environments to bright daylight. These sensors boast a dynamic range that reaches 140 $dB$, far surpassing the 60 $dB$ typically seen in high-quality frame-based cameras. 
A range of 140 $dB$ indicates the ability to handle brightness differences of up to 10,000,000:1, compared to only 1,000:1 for 60 $dB$. By minimizing saturation and preserving fine details, event cameras support applications like face detection or pose estimation under challenging lighting conditions.

\textit{Low Power Consumption}, as event cameras significantly reduce the amount of data produced by only capturing changes in the scene,  rather than full images as conventional frame-based cameras do. This feature makes event cameras ideal for long-term monitoring systems or wearable devices where power constraints are critical. For example, \cite{barchid2023spikingfer} demonstrated that Spiking-FER, when combined with event data, is 47.42$\times$ to 65.39$\times$ more energy-efficient than comparable artificial neural networks, highlighting the energy-saving potential of event-based systems and their suitability for low-power applications on edge devices.

\subsubsection{Privacy Preservation}
Another important characteristic that has brought event cameras to the spotlight is their potential for preserving user's privacy. Since they capture only dynamic scene changes, raw event data are inherently challenging to interpret compared to conventional RGB imagery. This feature adds a level of privacy by design, making event streams less likely to reveal sensitive identity information (\cite{becattini2024neuromorphicfaceanalysissurvey, DBLP:phd/ethos/AlObaidi20, Delilovic2021, Dong2023, Han2023}).
However, the assumption that event data are inherently privacy-preserving has been challenged by advancements in deep learning-based event-to-image reconstruction techniques (\cite{rebecq2019highspeedhighdynamic}), which can recover intensity map images from event streams and expose personal identity information. This has led to increased efforts to enhance the privacy of event-based data. In this direction, \cite{du2021event} proposed a 2D chaotic mapping-based algorithm that scrambles event positions and flips polarities, combined with a dynamic key-updating mechanism, ensuring data security while maintaining high efficiency on resource-constrained devices. Similarly, \cite{Zhang_2024} introduced an encryption framework to secure event streams during transmission, effectively preventing a direct application of computer vision models on the encrypted data. 
In the same line of research, \cite{ahmad2023personreidentificationidentificationevent, ahmad2024} formulated an anonymization strategy that randomizes event streams, making them unintelligible to human observers and demonstrating strong resilience against image reconstruction attacks, inversion, and adversarial learning attempts, while still retaining the information necessary for downstream tasks like person re-identification or human pose estimation.
\cite{bendig2024anonynoiseanonymizingeventdata} designed a novel pipeline for anonymizing event camera data by employing a learnable data-dependent noise prediction network combined with adversarial training, which was able to remove personally identifiable features to prevent re-identification.

When discussing privacy in the context of event cameras, it is crucial to consider it from the machine’s perspective, as the primary threat often arises from how machines interpret and utilize data. In their natural form, event data are completely unreadable to humans, appearing as sparse, asynchronous events that lack any recognizable visual information. However, when these events are reconstructed into frames, privacy concerns become more apparent for humans but are significantly greater for machines. Reconstructed frames are typically grayscale and of low resolution but retain substantial information due to their high temporal resolution, often reconstructed at rates approaching 5000 FPS. From a human viewpoint, these frames may appear inferior to traditional RGB images, especially in applications like action recognition where facial details are unclear. Yet, for machines, reconstructed frames hold significant value as they exploit motion edges and spatiotemporal patterns rather than visual clarity, leveraging the rich temporal data inherent in event-based recordings. This distinction emphasizes the need to develop privacy measures based on the machine’s capacity to extract sensitive information, recognizing that what seems visually obscure to humans may still be highly informative for automated systems.

\subsection{Challenges}
Despite their advantages, event cameras also present distinct challenges: while their spatially sparse and temporally asynchronous output allows for more efficient data storage, it necessitates the development of specialized algorithms to process and extract meaningful information that can be used in order to successfully downstream learning tasks. 
Handling events in an effective manner requires either to employ specialized frameworks, \textit{i.e.,} spiking neural networks (\cite{ghosh2009spiking}), or to represent the event data with more conventional formats, typically in the form of frames (see Section \ref{Sec:Data Repr}).
Traditional computer vision techniques are fundamentally designed for dense and synchronous images and therefore are not directly compatible with this novel data format (\cite{Gallego_2022}). When these methods, trained on RGB frames, are tested with input data coming from event cameras, they predictably struggle to perform well. The lack of a continuous flow of frames leads to a discrepancy with the underlying working assumptions that these methods are based on, resulting in low confidence or inaccurate results, as demonstrated in the study conducted by \cite{becattini2024neuromorphicfaceanalysissurvey} in the specific context of face detection and landmark prediction.
Furthermore, event-based sensors might exhibit inherent noise and non-idealities due to hardware constraints or environmental conditions that obscures data and complicates the interpretation. In this sense, the use of robust preprocessing techniques is essential to ensure a reliable performance.

\section{Event Camera Design and Processing} \label{Sec:Event Data}

In this section, we provide a comprehensive overview of event data, beginning with its representation techniques to facilitate efficient processing and compatibility with existing architectures. We then summarize and categorize the available datasets into real and synthetic, as well as body- and face-focused datasets. To address the scarcity of datasets, we further discuss event data simulators as a vital tool for generating synthetic data. Finally, we introduce event data compression, emphasizing its importance for real-time, human-centered applications of event cameras.

\subsection{Camera Models and Selection Criteria} \label{Sec:Cam_models}

The first Dynamic Vision Sensor (DVS) was introduced in 2008 by iniVation as the DVS128, offering a resolution of 128$\times$128 pixels \cite{lichtsteiner2008asynchronous}. Since then, several event camera models have been developed, primarily by companies such as iniVation\footnote{\url{https://inivation.com}}, Prophesee\footnote{\url{https://www.prophesee.ai}}, Samsung (\cite{samsung}), and CelePixel (\cite{celex}).
IniVation’s DAVIS series, such as the DAVIS240 and DAVIS346, combines event-based and frame-based sensing, offering resolutions up to 346$\times$260 pixels and dynamic ranges of 120 dB, making them versatile for mixed sensing tasks. Prophesee’s Gen3 and Gen4 cameras, with resolutions as high as 1280$\times$720 pixels and dynamic ranges exceeding 120 dB, are well-suited for applications requiring high spatial detail. CelePixel’s CeleX cameras provide features such as grayscale output and IMU integration, while Samsung’s DVS-GEN3 and DVS-GEN4 stand out with bandwidth capacities up to 1066 Meps (million events per second) for high-speed applications.

\cite{Gallego_2022} provided a comprehensive comparison of commercial and prototype event cameras serving as a critical reference for researchers to match camera capabilities with their specific application requirements.
When selecting an event camera for human-centered applications, specific criteria play a critical role and depend heavily on the task. For action recognition and human tracking, high temporal resolution and low latency (e.g., iniVation’s DAVIS240 at 12$\mu$s) are critical to capture fast motion dynamics. For facial analysis or anonymization, a higher spatial resolution is often more important to capture fine-grained details, as seen in the CeleX-IV (768$\times$640 pixels) or Prophesee Gen4 CD (1280$\times$720 pixels). Applications requiring operation in challenging lighting conditions, such as outdoor crowd density estimation, benefit from models with a high dynamic range (e.g., 120 dB in DAVIS346 or 143 dB in CeleX-IV). Additionally, power consumption is significant for wearable or mobile systems, where models like iniVation’s DAVIS240 (5-14 mW) are advantageous.

\subsection{Data Representation} \label{Sec:Data Repr}

As presented in the previous section, event cameras operate fundamentally differently from traditional frame-based cameras, resulting in asynchronous event streams encoding changes in intensity at each pixel with microsecond precision. These event streams, while rich in spatiotemporal information, require specialized processing techniques to extract meaningful features. Over time, various representations have emerged, each tailored to address specific challenges and applications. These representations can be categorized into the following groups based on their methodological approach and functional focus:

(1) Foundational representations: These include the earlier approaches such as Event Count (\cite{zhu2018evflownet}), Event Histogram, also referred to as event intensity frame, (\cite{liu2018adaptive}), Temporal Binary Representation (\cite{innocenti2021temporal}), Time Surface (\cite{lagorce2017hots}), and Memory Surface (\cite{pradhan2019nhar}) which prioritize simplicity and provide a quick way to interpret event data. \\

(2) Structural representations: These methods leverage advanced processing techniques to represent the spatial and temporal relationships of events. This includes the graph representation first proposed by \cite{bi2019graphobject} and \cite{bi2020graphtemporal} and then utilized as well by \cite{schaefer2022aegnn} and \cite{deng2022voxel}, which leverages graph theory to process event data both spatially and temporally. Similarly, \cite{tavanaei2019deep} proposed the Spiking tensor representation that tries to mimic the brain neurons as much as possible and represents event data as binary tensors. Moreover, Voxel Grid Representation - first proposed in \cite{zhu2018unsupervised} -  provides a more detailed approach by discretizing the event stream into 3D spatiotemporal grids. This representation is actually used to train complex networks such as video-based transformers and image reconstruction models.\\

(3) Reconstructed Frame Representations: In attempts to bridge the gap between event-based and frame-based frameworks, it also became popular in research to rely on representations like E2VID Frames (\cite{rebecq2019highspeedhighdynamic}) which allow us to mimic video frames and leverage the power of traditional Convolutional Neural Networks (CNNs) and even achieve better results in downstream applications compared to directly using event data.\\

(4) Fused representations:  Recently proposed by \cite{gao2023action}, the Learnable Multi-Fused Representation (LMFR) integrates multiple event representations, such as Time Surface, Event Frames, and Event Count, into a single embedding in a learnable manner. By leveraging their complementary features, LMFR enhances performance in complex tasks.\\

 Many other representations exist, but the most commonly used benchmarks are detailed in Table~\ref{tab:event_data_representations_updated}. Moreover, to provide a better understanding and facilitate comparison, we visualize a selection of these representations in Figure~\ref{fig:subfigures} using event data from the DVSGesture128 dataset, specifically for a hand-wave gesture. These visualizations highlight the diversity in how event data can be processed and interpreted for different applications. Note that for the graph representation in Figure~\ref{fig:subfig4}, while it is typically a 3D structure (like a point cloud) with events as nodes, for visualization purposes, we project it onto the temporal axis and represent it as a 2D structure.

While we have explored the different event data representations, it is equally important to evaluate their respective advantages and limitations to understand their suitability for various applications. First, Foundational representations, such as Event Count and Time Surface, are fast to compute and highly efficient, and can serve as lightweight baselines; however, their simplicity often results in significant loss of spatial information.  In contrast, structural representations, such as graph and spiking tensor approaches, capture spatio-temporal relationships effectively. Graph representations maintain connections between event nodes, making them effective for spatially complex tasks, though they can become computationally expensive in scenarios with dense movements or high event rates, often requiring filtering. Spiking tensors, on the other hand, align naturally with the neuromorphic nature of event cameras and are compatible with Spiking Neural Networks. However, SNNs face challenges like gradient optimization issues and remain less mature compared to CNNs. Similarly, reconstructed representations, such as intensity frames from E2VID, leverage the high temporal resolution of event cameras and leverage the power of the use of well-established CNN architectures; however, they often lose key spatial details due to their grayscale and low-resolution frames. Moreover, these frames can reintroduce redundancy through the huge amounts of frames generated and the static background information, which contradicts one of the main benefits of event-based systems. Finally, despite enhancing performance across tasks, the fused representation which combines multiple approaches si considered computationally intensive and requires significant resources to optimize effectively. Ultimately, the choice of representation depends on the specific application requirements, balancing computational efficiency, spatio-temporal complexity, and compatibility with existing architectures.

\subsection{Neural Network Architectures for Event Data}

In the literature, several prominent neural network architectures are utilized to process event data and downstream learning tasks, each leveraging unique paradigms. In this section, we outline the most commonly used approaches, briefly describing the foundational concepts.

(1) Spiking Neural Networks:
Event data are naturally compatible with SNNs, as they also operate based on an event-driven strategy. SNNs use discrete spikes rather than continuous activations, and perform well in handling spatio-temporal data while offering remarkable energy efficiency. Several works demonstrated how SNNs can better model the asynchronous nature of event data (\cite{liu2021event, barchid2023spikingfer}), while others further refined their application to real-world problems (\cite{bulzomi2023end, vicente2025spiking}).


(2) Graph Neural Networks (GNNs): 
GNNs are specialized artificial neural networks designed to process and analyze input data as graphs, \textit{i.e.,} structures that represent the relations (edges) between a collection of entities (nodes). GNNs directly operate on the spatio-temporal graph structure from the raw events, where the nodes can represent the event pixels and the edges the spatio-temporal dependencies between the nodes (\cite{wang2021event, gao2024hypergraph}).

(3) Convolutional Neural Networks:
CNNs are particularly suited for grid-structured data, \textit{i.e.,} images,  and aim to perform local feature extraction through convolution operations. To harness the capabilities of CNNs for event data, these data are either encoded into frame-based representations or processed by incorporating spatio-temporal convolutional layer in the architecture. The effectiveness of using CNNs in event data processing has been well-established(\cite{li2019lip, banerjee2022gaze, becattini2022facial}), allowing future studies to build upon these assumptions and further enhance event feature extraction (\cite{kanamaru2023isolated, goyal2023moveenet}).

(4) Transformers: 
Transformers moved away from traditional Recurrent Neural Networks (RNNs) and CNNs structures and revolutionized sequence-based data processing with their self-attention mechanism, which enables the allocation of higher weights to more significant information in the data. Depending on the tokenization strategy, events can be either processed in the form of frames, or in their raw format. Transformers allow for enhanced spatiotemporal feature extraction, effectively capturing the fine-grained dynamics of these types of data (\cite{cultrera2024spatio, zou2023eventpose}.

Each one of these architectures offers specific advantages, and the choice of which one to use is based on the specific characteristics of the event data and the requirements of the application. Table \ref{tab:network_classification} provides a comprehensive classification of key works that use these architectures, showcasing the diversity of methodologies and their applications.

\subsection{Event Simulators}

As highlighted in Figure \ref{fig:1}, event cameras are increasingly adopted in new domains. Initially applied in driver monitoring and robotics, their use has expanded to human motion analysis, including gait and action recognition, and more recently to face biometrics, capturing subtle facial movements. However, the widespread adoption of event cameras is hindered by the lack of publicly available datasets. This limitation has driven the development of event camera simulators, which convert conventional RGB video data into synthetic event streams by replicating the characteristics of event data as accurately as possible.

The first notable simulator, ESIM, was introduced by \cite{Rebecq18corl}, providing a foundational framework for generating synthetic event streams. \cite{gehrig2020video} build upon this work and developed the Vid2E simulator by adding an upsampling step to RGB videos, enhancing the accuracy and the ability of models trained with synthetic data to generalize for real data. \cite{hu2021v2e} later introduced V2E, a versatile simulator capable of producing raw event streams alongside grayscale frames and corresponding text files, broadening its applicability. \cite{lin2022dvsvoltmeter} proposed the DVS-Voltmeter, which improves synthetic event data quality by modeling the behavior of the DVS sensor using a unified approach that incorporates its circuit properties. Most recently, \cite{zhang2024v2ce} proposed the V2CE simulator, which stands out as the most precise event data simulator to date. Numerous other simulators have been developed (\cite{prophesee2023video, joubert2021event, han2024physical, mueggler2017event, carla2023dvs}); however, to the best of our knowledge, the ones discussed here are the most widely adopted and publicly available tools in the research community for generating synthetic event data.

The main advantage of event simulators is that they that they provide researchers with a means to bypass the need for expensive hardware and complex data collection during early experimentation for validating hypotheses, and also, to train simple networks for preprocessing tasks, like detecting regions of interest in real event data (\cite{becattini2022facial,barchid2023spikingfer}). In addition, simulators overcome one of the main challenges of working with real event data: synchronization between RGB and event data, particularly in applications where the training data requires both modality pairs \cite{berlincioni2023neuromorphic}).

However, despite their advantages, event simulators have notable drawbacks. First, they often fail to fully replicate the complexity and noise of real-world event data, which makes it harder for models trained on synthetic datasets to generalize effectively to real-world scenarios. Second, the quality of synthetic event streams heavily depends on the input RGB videos; if the videos lack high resolution and frame rate, the resulting event streams may miss critical details and temporal accuracy. Most critically, simulators inherently lose the high temporal resolution of event cameras, as they convert frame-based video inputs into events, reducing temporal precision from microseconds to frames per second, which can significantly affect tasks requiring fine-grained temporal information.

\subsection{Datasets}

The development and evaluation of event-based systems for human-centered applications heavily rely on publicly available datasets. In this survey, we categorize the datasets into two main groups based on their focus: body application datasets, which are designed for tasks such as action recognition, gait analysis, and human tracking, and face application datasets, which target applications like face detection, facial expression recognition, and anonymization. Each dataset is characterized by its unique properties, which we summarize in Tables \ref{tab:real_datasets} and \ref{tab:synthetic_datasets}.

\subsubsection{Real Datasets}
Table~\ref{tab:real_datasets} summarizes the human-centered real event datasets, focusing on body and face applications respectively. Each section of the table provides detailed information about the datasets, including the year of publication, the dataset name, the number of videos, the number of participants, and other information that could be relative to the corresponding application.

For body applications, key datasets include "Action Dataset TUM" for action recognition and "DVS128-Gait-Day" for gait analysis. Table~\ref{tab:real_datasets} also highlights newer datasets like "DailyDVS-200" and "THU-MV-E-ACT-50," which include multimodal data and a large number of classes. As for Face applications, notable datasets include "DVS-Lip" for lip reading and "NEFER" for Micro-Expression Recognition (MER). The more recent "VETEX" dataset combines multimodal data, including RGB and event streams, to enhance facial analysis tasks.

\subsubsection{Simulated Datasets}

Building on the challenges associated with real datasets—such as synchronization issues, noise, and limited diversity—we summarize in Table~\ref{tab:synthetic_datasets} the available synthetic datasets, categorizing them by their focus on body or face-related tasks. These datasets are generated in controlled environments, offering precise annotations and diverse scenarios that complement real-world data.

Unlike body applications, there are fewer publicly available synthetic datasets for face analysis. The currently available datasets are typically generated from the same benchmark RGB datasets, such as e-CK+ and e-MMI, and have been simulated multiple times by different researchers using tools like V2E, often with different parameters (\cite{Verschae2023EventGestureFacial, barchid2023spikingfer}). As a result, the generated event data cannot be published as standalone datasets but are typically shared as simulation code. In cases like NEFER dataset, \cite{berlincioni2023neuromorphic} utilized an unpublished simulated event dataset to train a face detector for creating synchronized bounding boxes, enabling event-based face detection on real data, even though the primary goal of the work was micro-expression recognition where a real dataset was collected. This has resulted in limited diversity in synthetic datasets for face-related tasks compared to body applications.

\subsection{Data Compression}

Despite event cameras being energy efficient, one key challenge is the significant data volume they generate, especially in real-time applications like robotics and video surveillance, where embedded systems require efficient storage and processing. While the asynchronous nature of event data reduces redundancy compared to traditional video streams, the sheer volume of events captured during high-speed motion or complex scenes remains a bottleneck. Several works have proposed compression techniques to address this challenge, leveraging both traditional and deep-learning-based methods (\cite{sezavar2024low, wang2023compress}).

Approaches to compress event data generally fall into multiple categories. Traditional methods include transforming events into frame-like representations for compatibility with standard video coding techniques (\cite{schiopu2022lossless}). Other methods, such as Spike coding, directly leverages the sparse and asynchronous nature of event streams to encode only significant changes, effectively reducing data size while preserving critical temporal information (\cite{sengupta2017spiketime, bi2018spike}). More recent research explores geometric-based structures as proposed by \cite{martini2022lossless} where they introduce a point cloud-based compression method capable of both lossy and lossless operations, achieving efficient data reduction.  \cite{huang2023evaluation} worked on point-cloud compression, further demonstrating that compression at high ratios maintains performance for tasks such as object detection and image reconstruction. They were able to achieve a compression ratio of 5 with lossless point cloud coding and with zero accuracy degradation on recognition tasks. Some large companies such as Google also contributed to that domain. In particular, Google designed \textit{Draco} which further extended previous methods by supporting additional attributes such as polarity, making it well-suited for event point clouds and stands out for its faster processing compared to other methods.
Recent advancements also explore deep-learning-based solutions. \cite{nguyen2021learning} proposed VoxelDNN, a model that captures the geometric structure of event data through convolutional networks, achieving both high compression efficiency and preservation of critical information. 

The growing importance of event data compression is further highlighted by recent initiatives such as the JPEG XE standard, developed by the JPEG committee (\cite{brites2024neuromorphic}). JPEG XE focuses on creating a standardized framework for efficiently representing event-based vision data, ensuring interoperability between sensing, storage, and processing systems.  This initiative reflects the increasing interest in event cameras within industry and research, as standardization efforts like these are critical for facilitating broader adoption. By targeting machine vision applications, this initiative addresses the unique challenges of event cameras, such as their sparse and asynchronous nature, while emphasizing their potential for real-world application. 

The techniques presented above, collectively enable event cameras to manage large-scale data effectively, facilitating their integration into real-time systems while maintaining the benefits of event-based vision.

\section{Applications} \label{Sec:Applciations}

In this section, we discuss the human-centered state-of-the-art applications of event data, divided into two macro areas: body and face. Table~\ref{tab:summary_application} summarizes the applications addressed in the literature using event cameras, referencing the corresponding state-of-the-art works. It is important to note that some papers tackle more than one application, and thus a reference may appear in multiple categories. In each subsection, we provide a more in-depth analysis of these research areas, explaining the relevant models for each application.

\subsection{Body}

In this subsection, we detail the applications of event-based data that require information from the full body of a person: gait recognition, action recognition, human tracking, and pose estimation.

\subsubsection{Gait Recognition}
One of the first human-centered applications of event-based camera, explored the feasibility of utilizing data obtained with this new sensor to address the classic problem of gait recognition. Gait recognition is a biometric technique aimed at identifying individuals based on their unique walking patterns. By mainly capturing motion with high temporal resolution and sparse data representation, researchers could effectively analyze and distinguish walking patterns to determine human identities. 

The first work on event-based gait recognition was presented in 2019 by \cite{wang2019ev}. 
Due to the noisy and asynchronous nature of events, traditional vision-based gait recognition algorithms were unsuitable for such data. To address this challenge, they proposed a novel approach called EV-Gait. This method leverages motion consistency to effectively reduce noise in event streams and employs a deep neural network to recognize gait patterns from the asynchronous and sparse event data, making it specifically tailored to the capabilities and challenges of this technology.

Over time, various architectures have been proposed to tackle the task of event-based gait recognition. An early work by \cite{sokolova2019human} introduced a pipeline composed of five consecutive steps: visualization of the event stream, human figure detection, optical flow estimation, human pose estimation, and finally, gait recognition based on neural features. This approach achieved performance comparable to conventional methods using color videos. Another approach by \cite{tao2024gaitspike} utilized SNNs to process event data, introducing a domain-specific Locomotion-Invariant Representation (LIR). LIR replaced the static Cartesian coordinates of the raw event camera data with a floating polar coordinate system centered on the motion axis, improving the representation's adaptability to dynamic scenarios. Further innovations in \cite{fu2023hypergraph} include the use of hypergraph neural networks for gait recognition. This method employed an event flow downsampling module to reduce data volume without compromising discriminability, an event feature extraction module to convert events into graph nodes, and a spatiotemporal hypergraph convolution module to construct a hypergraph, extract spatiotemporal features, and obtain pedestrian gait features.

Comparative works have also emerged in the literature. In 2022, \cite{eddine2022gait3} conducted experiments using a baseline algorithm based on gait energy images adapted to event-camera output. They compared this approach to results from RGB and thermal videos using the same algorithm, demonstrating a distinct advantage for event-based data. \cite{wang2021event} investigated different representations of event streams for deep neural network classifiers. They proposed novel event-based gait recognition approaches using two distinct representations: graph-based and image-like. These methods leveraged graph convolutional networks and convolutional neural networks, respectively, showcasing the versatility of event-based data for gait recognition.

\subsubsection{Action Recognition}
Action recognition is a major research focus in computer vision due to its importance in applications such as security and human-computer interaction (\cite{adra2024comparative}). Research in this field has advanced with the use of bio-inspired event sensors which capture only the activity in their field of view and automatically differentiate the foreground from the background, making them ideal for recognizing human actions.

 \cite{liu2021event} made an early attempt to apply motion information to event-based action recognition by extracting motion features from events, progressing from local to global perception. On the other hand, 
 \cite{ren2023spikepoint} introduced SpikePoint, a novel end-to-end point-based SNN architecture that processes event data as cloud data and converts them into spikes using rate coding. More recently, in 2025, 
 \cite{vicente2025spiking} demonstrated that spiking neurons can enable temporal feature extraction in feed-forward neural networks without requiring recurrent synapses, and how recurrent SNNs can achieve performance comparable to LSTMs with fewer parameters, validating their approach in action recognition. 

Beyond SNNs, other architectures have been explored to create more lightweight models. \cite{de2023eventtransact} proposed a video transformer-based framework that acquires spatial embeddings per event-frame and utilizes a temporal self-attention mechanism. This approach separates spatial and temporal operations, making the video transformer more computationally efficient than other video transformers. \cite{ren2023ttpoint} proposed a point cloud-based method for action recognition using event data, featuring a hierarchical structure that distinguishes local and global features. Their model is lightweight, thanks to the application of tensor decomposition to compress the data.

In more recent works, \cite{gao2023action} introduced EV-ACT, an event-based action recognition framework that uses a slow-fast network to fuse motion and appearance-related features. One of their key contributions is the Learnable Multi-Fused Representation, which integrates multiple event representations, such as time surfaces, event frames, and event count, into a single embedding. In an extension of their work, \cite{gao2024hypergraph} proposed HyperMV, a multi-view event-based action recognition framework utilizing hypergraphs and a hypergraph neural network to capture relationships across viewpoint and temporal features.

Additionally, \cite{plizzari2022e2} proposed two new strategies; directly processing event-camera data with traditional video-processing architectures and using event data to extract optical flow information. They also compared the performance of different pairings of event, RGB, and optical flow. Another comparative study was conducted by \cite{wang2025dailydvs}, where in addition to introducing their benchmark database, DailyDVS-200, they evaluated it using 12 event-based architectures for action recognition.

\subsubsection{Pose Estimation}
Human Pose Estimation refers to the identification of key body joints in a human and plays a vital role in many human-centered tasks (\cite{rafi2020self}). In fields like robotics, IoT, and smart home applications, pose estimation is the initial step that supports subsequent processes such as action recognition, posture analysis, and emotion and intent detection (\cite{goyal2023moveenet}).

In 2019, \cite{sokolova2019human} attempted the first pose estimation using event-based human data. Although their primary goal was human gait recognition, they also addressed several auxiliary challenges, such as moving object detection and human pose estimation in event-based video sequences. Their model focused on detecting areas of interest and subsequently computing optical flow to estimate the positions of key pose points. In more recent approaches, \cite{goyal2023moveenet} presented a system for high-frequency 2D human pose estimation for a single person. The core of their approach is the use of a lightweight, image-like event representation that resolves the issue of static body parts disappearing and allows pre-training on widely available frame-based datasets with high-accuracy ground truth, followed by fine-tuning on native event-camera datasets. 

\cite{zou2023eventpose} introduced the first end-to-end method for 3D human pose tracking using only event data, leveraging Spiking Neural Networks. In 2024, \cite{kohyama20243d} proposed a method that exclusively uses event data to create 3D voxel representations by moving an event camera around a stationary body \cite{kohyama20243d}. This method reconstructs human pose and mesh through attenuated rays while fitting statistical body models to preserve high-frequency details.

\subsubsection{Human Tracking}
In recent years, \cite{mitrokhin2018event} and \cite{ramesh2020tld} have proposed several approaches for event-based object tracking, primarily focusing on tracking objects with simple shapes. Building on this, a new research direction has emerged, addressing the relatively novel problem of tracking 3D human inputs solely based on event streams from an event camera, thereby completely eliminating the need for additional dense input images.
In 2023, \cite{eisl2023singlehumantracking} presented a novel framework for tracking humans using a single event camera, comprising three main components. First, a Graph Neural Network was trained to identify a person within the stream of events. To preserve the sparse nature of the event data and leverage its high temporal resolution, batches of events are represented as spatio-temporal graphs. Next, the person was localized in a weakly-supervised manner via Class Activation Maps to their graph-based classification model, eliminating the need for ground truth human positions during training followed by a Kalman filter for tracking.

Existing works in pose tracking either require the presence of additional grayscale images to establish a reliable initial pose as it is the case in \cite{xu2020eventcap} or disregard temporal dependencies altogether by collapsing segments of event streams into static event frames like in \cite{rudnev2021eventhands}. \cite{zou2023eventpose} introduced a dedicated end-to-end sparse deep learning approach for event-based 3D human pose tracking where the task is achieved without any reliance on frame-based images. Their method is based on a Spiking Neural Network, with the incorporation of a Spike-Element-Wise ResNet and a novel Spiking Spatiotemporal Transformer.

\subsection{Face}

In this subsection, we analyze the use of event-based data for tasks that involve solely the face of an individual. Those applications are face detection, identity recognition, lip-reading, eye blinking and gaze analysis and microexpression and emotion recognition.

\subsubsection{Face Detection}
An early application of event-based facial data was face detection, a task that involves identifying and locating human faces within an image or video stream. Face detection serves as a foundational step for various facial applications, including identity recognition, soft biometric estimation, and behavior analysis.

In 2016, \cite{barua2016direct} developed a pioneering face detection model based on translating event streams into large-scale images using a patch-based approach. Their method involved learning a sparse dictionary of patches to reconstruct both simulated and real event data, even in noisy conditions. Their event-based face detection framework achieved results comparable to the traditional Viola-Jones face detector (\cite{viola2001rapid}). \cite{bissarinova2023faces} proposed an architecture that utilizes events accumulated over time and incorporates past event information for effective face detection. They presented 12 models trained on their dataset to predict bounding boxes and facial landmark coordinates. Additionally, they showcased real-time face detection capabilities using event-based cameras and their models. More recently, \cite{himmi2024ms} defined the concept of multispectral events, capturing data across multiple spectral bands to enhance event-based face detection. They demonstrated that multispectral events significantly improve face detection performance compared to monochromatic grayscale events, surpassing even conventional multispectral image performance.

As face detection often precedes other facial processing tasks, several studies have combined face detection with additional applications. In 2020, \cite{lenz2020event} introduced the first purely event-based method for face detection, relying on eye-blink detection. They analyzed the temporal signature of eye blinks and employed a Gaussian tracker to statistically measure pixel activity in the event stream. In 2021, \cite{ryan2021real} proposed GR-YOLO, a novel neural network for face and eye detection using event cameras, specifically in driver monitoring systems. Their architecture, based on YOLOv3-tiny, incorporated a fully convolutional gated recurrent unit layer. By 2023, \cite{ryan2023real} extended this work by introducing a two-stage event-based multi-task facial analytics framework. The first stage used a CNN to locate and track faces and eyes, while the second stage employed another CNN to estimate head pose, eye gaze, and occlusions within a multi-task learning setup. Building on previous work, \cite{iddrisu2024evaluatingimagebasedfaceeye} utilized a Temporal Binary Representation of event data and trained a GR-YOLO model, comparing its performance to YOLOv8 for face and eye detection tasks.

\subsubsection{Identity Recognition}

Identity recognition via face images is a biometric technology that identifies or verifies individuals based on their distinct facial features. This task is critical for numerous applications requiring reliable verification or identification, as the face is a unique and easily accessible trait crucial for enhancing security systems. 
Identity recognition via face images is a biometric technology that identifies or verifies individuals based on their distinct facial features. This task is critical for numerous applications requiring reliable verification or identification, as the face is a unique and easily accessible trait crucial for enhancing security systems. So far in the literature, Identity recognition from event data has been performed with the help of other auxiliary tasks such as eye blink characterization or facial dynamics derived from speech.

In 2021, \cite{chen2020neurobiometric} proposed the first neuromorphic, event-based biometric authentication system. Their method for identity recognition relied on eye blink characterization. They defined a set of biometric features describing the motion, speed, energy, and frequency signals of eye blinks, leveraging the microsecond temporal resolution of event densities. Using these features, they trained both an ensemble model and a non-ensemble model with their NeuroBiometric dataset for biometric authentication. In a subsequent work, \cite{moreira2022neuromorphic} explored the potential of event sensors for identity recognition through a novel facial characteristic: facial dynamics derived from speech. They also validated the contribution of facial motion to human face identity categorization. Their approach involved aggregating events into frames, normalizing them, and grouping them into so-called "face tokens," which were then processed by a spatio-temporal 3D CNN to extract insights about the individual's identity. 

\subsubsection{Lip Reading}

Voice Activity Detection (VAD) is a technique used to identify and isolate segments of speech within an audio stream. Event cameras, with their high temporal resolution and ability to capture micro-movements, are particularly beneficial for this task. By accurately detecting subtle mouth movements, event data can enhance the precision of VAD, as well as related applications like lip-reading, where understanding spoken language relies on analyzing lip motions.

\cite{savran2018energy} explored for the first time voice activity detection (VAD) using event data. In their VAD pipeline, they leveraged event-based facial data by adding an initial module in their pipeline where lip activity was filtered spatio-temporally and then detected jointly through probabilistic estimation. In a later work, \cite{savran2023fully} continued their research proposing an event intensity-based method for VAD by designing a fully convolutional network to segment vocally active durations efficiently. In their approach, the raw event sequence was first processed to ensure that voice-related temporal information was preserved in a low-dimensional representation. Subsequently, a fully convolutional VAD network was constructed to carry out the detection task. In 2023, \cite{kanamaru2023isolated}  presented an event camera-based lip-reading method for isolated single-sound recognition. Their pipeline included imaging from event data, face and facial feature detection, and recognition using a Temporal Convolutional Network (TCN). Their findings demonstrated that event-based cameras achieved higher lip-reading accuracy than traditional frame-based cameras. Furthermore, the authors showed that combining two modalities, the frame-based camera and the event-based camera, yielded higher accuracy than using either modality alone. In the same yar, \cite{bulzomi2023end} proposed the first SNN model for event-based lip reading, achieving competitive results compared to state-of-the-art artificial neural networks.

An innovative approach by \cite{li2019lip} combined video and audio data for the first time. The authors introduced a lip-reading deep neural network that fused the asynchronous spiking outputs of two bio-inspired silicon multimodal sensors: the Dynamic Vision Sensor and the Dynamic Audio Sensor. Their classification process, based on event-based features generated from the spikes of these sensors, was tested on the GRID visual-audio lipreading dataset. Similarly, \cite{rios2023lipsfus}. utilized CNNs to process visual and auditory information in their self-collected dataset, which involved participants speaking a set of words. The visual information was derived from lip movements captured by event cameras as the subjects articulated words. The event activity was converted into histograms, which a CNN further processed.

\subsubsection{Eye Blinking and Gaze Analysis}

Eye movement has been extensively studied in the biometrics community due to its potential for applications in authentication, gaze tracking, and behavioral analysis. Indeed, when addressing eye-blinking characterization from event-based data, researchers often solve this problem as an auxiliary task for other major objectives.
\cite{lenz2020event} implemented a low-power human eye-blink detection method designed to exploit the high temporal precision provided by event-based cameras. Similarly, \cite{chen2020neurobiometric} developed an authentication system based on eye blinks captured with an event camera, achieving high accuracy with computationally simple processes.

A different area of focus is gaze and eye tracking. In 2022, \cite{angelopoulos2021event}, defined a pipeline for gaze tracking that combined frames recorded at a fixed sampling rate with asynchronous events capturing eye motion at high speed. Their method outputs a gaze point derived from an estimate of the pupil, forming an almost continuous tubular structure that outlined the pupil’s movement. \cite{banerjee2022gaze} proposed a novel event-encoding technique that converted motion event logs into six-channel images. They then designed a CNN to predict gaze using the encoded events from the event camera. In another study, \cite{iddrisu2024evaluatingimagebasedfaceeye} employed an event simulator to convert RGB videos into event-based data. Their approach involved accumulating events into binary frames and aggregating these frames into a single one to enhance the density and quality of the simulated data. They subsequently compared the performance of different state-of-the-art models using the generated event data.

In 2021, \cite{ryan2021real} leveraged event-based data to create a low-energy consumption model for simultaneously detecting and tracking faces and eyes, specifically for driver monitoring applications. They developed a customized fully convolutional neural network for this purpose. Later, in 2023, \cite{ryan2023real} extended their work for the same application by designing a multitask neural network for real-time facial analysis. This new model simultaneously estimated head pose, eye gaze, and facial occlusions. It was trained on synthetic data and evaluated in real-world scenarios.

\subsubsection{Micro-expressions}

Facial Emotion Recognition (FER) is a technology that analyzes facial expressions from static images and videos to infer a person's emotional state. Recent advancements in the FER domain have focused on estimating microexpressions, subtle and rapid facial movements often performed involuntarily, due to their strong connection with emotions as defined by the Facial Action Coding System.

\cite{becattini2022facial} pioneered the application of event cameras for FER using synthetic event data. Leveraging an event-camera simulator, they generated synthetic event streams and transferred face bounding boxes onto the data. Cropped face sequences were then processed by a CNN, followed by a long short-term memory network to account for the temporal dimension.
In 2023, \cite{barchid2023spikingfer} introduced "Spiking-FER," a deep convolutional SNN inspired by ResNet18, achieving superior performance compared to traditional visible-domain methods. \cite{berlincioni2023neuromorphic} classified microexpressions into three categories neutral, positive, and negative using a baseline 3D-CNN. Similarly, \cite{guo2023gleffn} proposed a lightweight approach utilizing a global-local event feature fusion network, which merged local count images with global dense optical flow to extract deeper features for FER.

In 2024, three studies further advanced the use of event data for microexpression estimation. \cite{xiao2024estme} developed a system with two key components: the Event-Enhanced Motion Extractor, which amplified subtle movements, and the Event-Guided Attention module, which focused on crucial facial regions for microexpression analysis. \cite{cultrera2024spatio} introduced the first video transformer model for action unit classification from event streams, significantly improving accuracy. Finally, \cite{adra2024beyond} conducted experiments on their novel dataset revealing that thermal and event-based modalities outperformed visible-spectrum cameras for microexpression recognition. Although thermal images provided the best performance under varying illumination conditions, event data also demonstrated strong capabilities, as its high temporal resolution proved more effective at capturing small facial movements than traditional RGB cameras.

\subsection{Discussion}

In this section, we have extensively reviewed the various applications of data obtained from neuromorphic cameras in human-centered contexts. Neuromorphic human analysis is a relatively new field of research. Nonetheless, several studies have highlighted the effectiveness of neuromorphic cameras for a variety of applications related to both the human body and face, offering notable advantages compared to traditional computer vision techniques. For instance, as demonstrated in Table~\ref{tab:rgb_vs_event}, neuromorphic cameras show significant improvements in tasks such as action and microexpresison recognition when compared to RGB-based methods.
However, for other tasks, the reported improvements are marginal or even negligible. For example, in gait recognition, the observed gains are minimal, and for applications like face detection and lip reading, the performances of neuromorphic cameras are often comparable to those achieved with RGB-based approaches. This suggests that, while event data holds promise, its benefits are not yet universally realized across all applications. Additionally, a critical limitation in the current state of research is the lack of standardized benchmark datasets. There is a tendency for researchers to report results on newly created databases, often without direct comparison to existing datasets, making it challenging to objectively evaluate and compare progress across studies.

Moreover, over the past decade, computer vision has made remarkable advancements in their AI-based architectures such as CNNs and vision transformers. These models are highly optimized to extract detailed and meaningful information from RGB data, resulting in state-of-the-art performances across a wide range of tasks while other modalities such as event or thermal data have not received the same level of attention in model development. Current neural networks are not inherently designed to fully leverage the unique characteristics of these modalities, which limits their potential. While event-based or thermal sensors may provide additional, task-specific information that could be more useful than RGB data in certain scenarios, the lack of tailored architectures results in RGB data often outperforming these modalities. This is evident in the treatment of event data, where, as noted in Table~\ref{tab:network_classification}, processing frequently involves converting event streams into representations that mimic the structure of RGB frames to enable their use with pre-existing CNN architectures.

\section{Conclusion and Future Perspectives} \label{Sec:Conclusions}

Human-centered applications are one of the foundations of computer vision research, addressing challenges and opportunities in diverse areas such as surveillance, biometric authentication, autonomous driving, and behavioral analysis. While traditional frame-based methods using RGB cameras have achieved remarkable advancements, their limitations in temporal resolution, motion blur, and low-light conditions have increased the interest in neuromorphic cameras. These sensors represent a paradigm shift, capturing asynchronous pixel intensity changes that provide high temporal resolution, robustness in challenging environments, and reduced computational demands.

This survey offers a comprehensive overview of the progress and potential of event-based cameras for human-centered applications. By categorizing advancements in human body- and face-related tasks, we highlight the progress made in recent years, emphasizing the strengths and innovations of architectures leveraging event-based data. Presenting the current state of the art, identifying challenges, and suggesting future directions, this survey aims to guide researchers in exploiting the potential of event-based cameras for human-centered applications. 

By analyzing the properties of event data, we observed how it is uniquely suited for machines due to characteristics such as high temporal resolution and low latency. These attributes provide AI models with richer and more precise information compared to traditional RGB data. However, event data is less intuitive and interpretable for humans, making it better aligned with AI capabilities than with human understanding.
Moreover, in our state-of-the-art review, we identified a significant drawback in many of the models presented: researchers often focus on demonstrating the suitability of event data for specific applications and few works conduct thorough comparisons with RGB-based methods. This lack of direct performance comparisons highlights the early developmental stage of event-based models, which in some cases have yet to reach the maturity required for widespread adoption.
Additionally, we want to remark that the future of acquisition sensors remains uncertain, particularly as generative AI continues to advance enabling the creation of highly accurate synthetic images that can be used to train high-precision networks without compromising individual privacy. Such synthetic datasets have the potential to complement or even replace real-world event-based data in certain scenarios.

However, looking ahead, several promising research directions emerge. First, there is a growing need for more robust and standardized event-based datasets, particularly those capturing real-world conditions across diverse human-centered scenarios. Second, the development of hybrid systems that integrate event-based and frame-based modalities offers significant potential, as these systems could enhance the strengths of both approaches. Finally, advances in event representation and processing methods, along with deep learning architectures specifically optimized for event data, are essential for unlocking the full potential of these sensors.



\section*{Conflict of Interest Statement}
The authors declare that the research was conducted in the absence of any commercial or financial relationships that could be construed as a potential conflict of interest.

\section*{Author Contributions}

MA, SM and NM contributed equally to the conception and design of the study and wrote the first draft of the manuscript. JLD
contributed to manuscript conception as well as its revision.

\section*{Funding}

This research is a part of the HEIMDALL project, funded by the BPI as part of the AAP I-Demo.
Additionally, the work was supported by the European Union’s Horizon Europe research and innovation program under Grant Agreement No 101094831 for the Converge-Telecommunications and Computer Vision Convergence Tools for Research Infrastructures project.





\bibliographystyle{Frontiers-Harvard} 
\bibliography{bibliography}




\newpage

\section*{Figures}

\setcounter{figure}{2}
\setcounter{subfigure}{0}
\begin{subfigure}
\setcounter{figure}{2}
\setcounter{subfigure}{0}
    \centering
    \begin{minipage}[b]{0.33\textwidth} 
        \includegraphics[width=\linewidth]{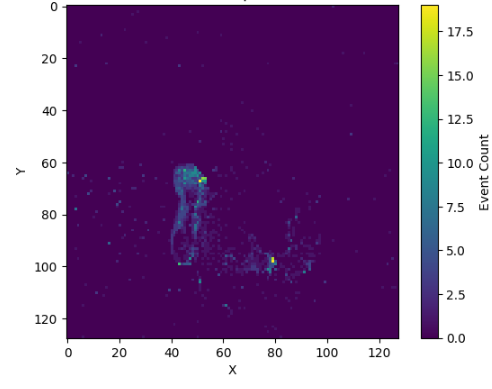}
        \caption{Event count }
        \label{fig:subfig1}
    \end{minipage}  
    \begin{minipage}[b]{0.33\textwidth} 
        \includegraphics[width=\linewidth]{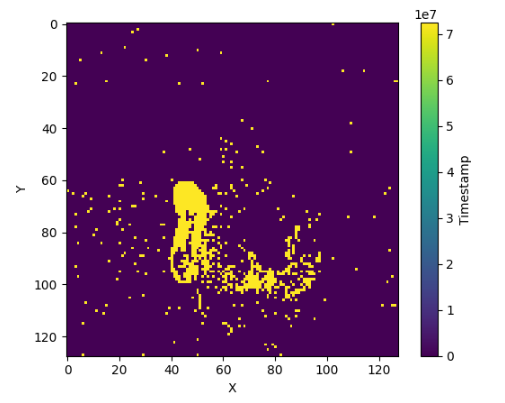}
        \caption{Time surface}
        \label{fig:subfig2}
    \end{minipage}
    
    \vspace{0.5cm} 

    \begin{minipage}[b]{0.33\textwidth} 
        \includegraphics[width=\linewidth]{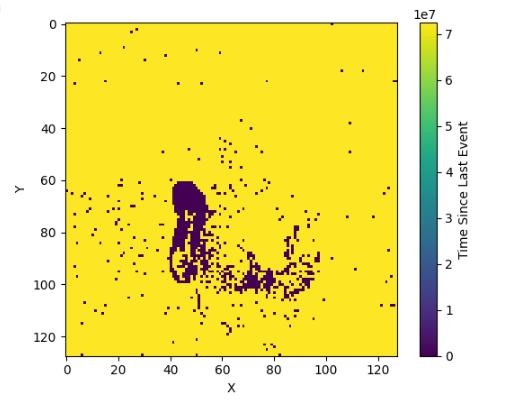}
        \caption{Memory surface}
        \label{fig:subfig4}
    \end{minipage}
    \begin{minipage}[b]{0.33\textwidth} 
        \includegraphics[width=0.95\linewidth]{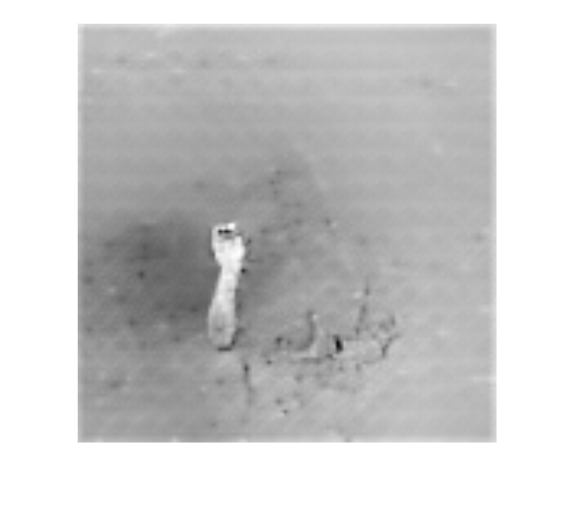}
        \caption{Frame reconstruction}
        \label{fig:subfig5}
    \end{minipage}

    \vspace{0.5cm} 

    \begin{minipage}[b]{0.33\textwidth} 
        \includegraphics[width=\linewidth]{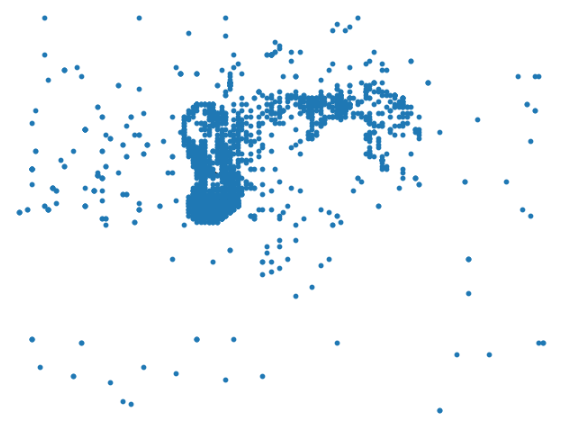}
        \caption{Graph }
        \label{fig:subfig4}
    \end{minipage}
    \begin{minipage}[b]{0.33\textwidth} 
        \includegraphics[width=\linewidth]{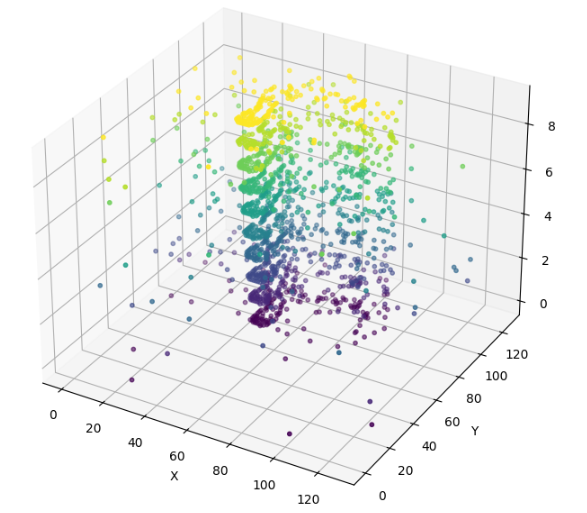}
        \caption{Voxel grid }
        \label{fig:subfig5}
    \end{minipage}

    \setcounter{figure}{2}
    \setcounter{subfigure}{-1}
    \caption{ Visualizations of various event data representations derived from the DVSGesture128 dataset, specifically illustrating a hand-wave gesture.}
    \label{fig:subfigures}
\end{subfigure}

\clearpage

\section*{Tables}

\begin{table}[h]
\footnotesize
\centering
\caption{The table presents applications of event cameras for human data along with an exhaustive selection of relevant works for each category. The applications are categorized into two main areas: face and body.
\vspace{0.5cm}}
\label{tab:summary_application}
\begin{tabular}{ccccc}
\multicolumn{4}{c}{\textbf{Body}}                                               & \textbf{}                                                                        \\ \cline{1-4}
\multicolumn{1}{c|}{\begin{tabular}[c]{@{}c@{}}Human\\ Tracking\end{tabular}} & \multicolumn{1}{c|}{\begin{tabular}[c]{@{}c@{}}Gait\\ Recognition\end{tabular}}     & \multicolumn{1}{c|}{\begin{tabular}[c]{@{}c@{}}Action\\ Recognition\end{tabular}} & \begin{tabular}[c]{@{}c@{}}Pose\\ Estimation\end{tabular}    &   \\ \cline{1-4}

\multicolumn{1}{c|}{\begin{tabular}[c]{@{}c@{}}\cite{eisl2023singlehumantracking}\\ \cite{xu2020eventcap}\end{tabular}} 

& \multicolumn{1}{c|}{\begin{tabular}[c]{@{}c@{}}\cite{wang2019ev}\\ \cite{sokolova2019human}\\ \cite{wang2021event}\\ \cite{eddine2022gait3}\\ \cite{fu2023hypergraph}\\ \cite{tao2024gaitspike}\end{tabular}}   

& \multicolumn{1}{c|}{\begin{tabular}[c]{@{}c@{}} \cite{liu2021event} \\
\cite{plizzari2022e2} \\
\cite{ren2023ttpoint} \\
\cite{ren2023spikepoint} \\
\cite{de2023eventtransact} \\
\cite{gao2023action} \\
\cite{gao2024hypergraph} \\
\cite{vicente2025spiking} \\
\cite{wang2025dailydvs} 
 \end{tabular}}    

& \begin{tabular}[c]{@{}c@{}}\cite{sokolova2019human} \\ \cite{zou2023eventpose} \\ \cite{goyal2023moveenet} \\ \cite{kohyama20243d}\end{tabular}       
&   \\

\multicolumn{1}{l}{}   & \multicolumn{1}{l}{}       & \multicolumn{1}{l}{}    & \multicolumn{1}{l}{}        & \multicolumn{1}{l}{}     \\
\multicolumn{5}{c}{\textbf{Face}}   \\ \hline

\multicolumn{1}{c|}{\begin{tabular}[c]{@{}c@{}}Face\\ Detection\end{tabular}} & \multicolumn{1}{c|}{\begin{tabular}[c]{@{}c@{}}Identity\\ Recognition\end{tabular}} & \multicolumn{1}{c|}{\begin{tabular}[c]{@{}c@{}}Lip\\ Reading\end{tabular}}        & \multicolumn{1}{c|}{\begin{tabular}[c]{@{}c@{}}Eye Blinking\\ \& Gaze\end{tabular}} & \begin{tabular}[c]{@{}c@{}}Microexpression \&\\ Emotion Recognition\end{tabular} \\ \hline

\multicolumn{1}{c|}{\begin{tabular}[c]{@{}c@{}}\cite{barua2016direct} \\ \cite{lenz2020event} \\ \cite{ryan2021real} \\ \cite{bissarinova2023faces} \\ \cite{ryan2023real} \\ \cite{himmi2024ms} \\ \cite{iddrisu2024evaluatingimagebasedfaceeye}
 \end{tabular}}   

& \multicolumn{1}{c|}{\begin{tabular}[c]{@{}c@{}} \cite{chen2020neurobiometric}\\ \cite{moreira2022neuromorphic}\end{tabular}} 

    & \multicolumn{1}{c|}{\begin{tabular}[c]{@{}c@{}}\cite{savran2018energy}\\ \cite{li2019lip}\\  \cite{rios2023lipsfus} \\ \cite{savran2023fully} \\ \cite{kanamaru2023isolated} \\ \cite{bulzomi2023end}\end{tabular}}

& \multicolumn{1}{c|}{\begin{tabular}[c]{@{}c@{}} \cite{lenz2020event} \\ \cite{chen2020neurobiometric} \\ \cite{angelopoulos2021event} \\ \cite{ryan2021real} \\ \cite{banerjee2022gaze} \\ \cite{ryan2023real} \\ \cite{iddrisu2024evaluatingimagebasedfaceeye}\end{tabular}} 
& \begin{tabular}[c]{@{}c@{}}\cite{becattini2022facial} \\ \cite{barchid2023spikingfer} \\ \cite{berlincioni2023neuromorphic} \\ \cite{guo2023gleffn} \\ \cite{xiao2024estme} \\ \cite{cultrera2024spatio} \\ \cite{adra2024beyond} \end{tabular}                                 
\end{tabular}
\end{table}

\begin{table}[]
\centering
\footnotesize

\caption{Papers presented in this survey, classified by the type of AI architecture used for their models}
\vspace{0.5cm}

\label{tab:network_classification}
\begin{tabular}{|c|c|c|c|c|}
\hline
\textbf{SNN}    & \textbf{Graph NN}   & \textbf{CNN}      & \textbf{Transformers}   & \textbf{Not AI-based}  \\ \hline

\begin{tabular}[c]{@{}c@{}} \cite{liu2021event}\\  \cite{barchid2023spikingfer} \\ \cite{ren2023spikepoint}\\ \cite{bulzomi2023end} \\ \cite{tao2024gaitspike}\\ \cite{vicente2025spiking}\end{tabular} & 

\begin{tabular}[c]{@{}c@{}} \cite{wang2021event} \\ \cite{eisl2023singlehumantracking}\\ \cite{fu2023hypergraph}\\ \cite{gao2024hypergraph} \end{tabular} &

\begin{tabular}[c]{@{}c@{}}\cite{li2019lip} \\ \cite{wang2019ev}\\ \cite{sokolova2019human}\\ \cite{ryan2021real}\\ \cite{banerjee2022gaze} \\ \cite{becattini2022facial} \\ \cite{moreira2022neuromorphic} \\ \cite{plizzari2022e2}\\ \cite{ryan2023real} \\ \cite{gao2023action}\\ \cite{rios2023lipsfus} \\ \cite{bissarinova2023faces} \\ \cite{berlincioni2023neuromorphic} \\ \cite{goyal2023moveenet} \\ \cite{kanamaru2023isolated} \\ \cite{xiao2024estme} \\ \cite{kohyama20243d} \\ \cite{adra2024beyond} \\ \cite{iddrisu2024evaluatingimagebasedfaceeye} \end{tabular} &

\begin{tabular}[c]{@{}c@{}}\cite{xu2020eventcap}\\  \cite{de2023eventtransact}\\ \cite{zou2023eventpose} \\ \cite{cultrera2024spatio}\end{tabular} & 

\begin{tabular}[c]{@{}c@{}}\cite{barua2016direct}\\ \cite{savran2018energy} \\ \cite{lenz2020event} \\ \cite{chen2020neurobiometric} \\ \cite{angelopoulos2021event} \\ \cite{eddine2022gait3}\\ \cite{ren2023ttpoint} \\ \cite{guo2023gleffn} \\ \cite{savran2023fully} \\ \cite{himmi2024ms} \end{tabular} \\ \hline
\end{tabular}
\end{table}


\begin{table}[]
\footnotesize
\centering
\caption{Event Data Representations and Their Details}
\vspace{0.5cm}
\renewcommand{\arraystretch}{1.5} 
\begin{tabular}{|l|p{12cm}|} 
\hline
\textbf{Name} & \textbf{Details} \\ \hline
Event Count & Event data is aggregated by counting the number of events that occur at each pixel within a fixed time interval. This approach provides a straightforward summary of activity, often used as a baseline representation. \\ \hline
Event Histogram & Similar to the event count, but instead of a single time interval, events are grouped and counted in temporal bins, creating a distribution of event activity that captures variations with more levels of detail. \\ \hline
Time Surface / Surface of Active Events & Represents data as a continuous map where each pixel value corresponds to the most recent timestamp of an event at that location. This highlights recent activity and is often used to track motion or identify edges. \\ \hline
Memory Surface & Event data are represented as a temporal map where each pixel's value indicates the time elapsed since the last event occurred at that location within a fixed time window. This approach encodes temporal information by retaining a "memory" of inactivity, making it useful for identifying patterns, and tracking regions with recent or ongoing motion. \\ \hline
Voxel Grid & Event data is sliced temporally into small time intervals, creating a sequence of event slices. These slices are then stacked into a 3D grid, where each voxel represents the activity in a spatial region during a specific time window. This allows for preserving both spatial and temporal resolution. \\ \hline
Spike Tensor & Represents data as binary tensors indicating the occurrence of spikes in specific spatiotemporal locations. The tensor is separated into two channels for positive and negative polarities. \\ \hline
Graph & Represents data as a graph, where events are treated as nodes in a graph with polarity as the node feature. Then, edges are created between nodes to represent spatiotemporal relationships, often used for tasks like pattern recognition. \\ \hline
E2VID Frame & Represents data as reconstructed frames by using neural networks to convert the sparse event stream into intensity frames. This allows event data to be used with traditional frame-based computer vision methods. \\ \hline
Temporal Binary Representation & Events are first stacked together into intermediate binary representations where each pixel can be considered as a binary string. These frames are then grouped into a single frame by applying binary to decimal conversion. Most popular in face analysis applications. \\ \hline
\end{tabular}
\label{tab:event_data_representations_updated}
\end{table}

\begin{table}[]
\footnotesize
\centering
\caption{Real event-based datasets for human-centered applications.}
\vspace{0.5cm}
\renewcommand{\arraystretch}{1.5}

\begin{tabular}{cccccccc}
\multicolumn{8}{c}{\textbf{Body Datasets}} \\ \hline
\textbf{Year} & \textbf{Authors} & \textbf{Name} & \textbf{\# Videos} & \textbf{\# People} & \textbf{Modalities} & \textbf{Application} & \textbf{\# Classes} \\ \hline 
2019 & \cite{miao2019neuromorphic}      & Action Dataset TUM      & 291     & 15     & EV       & Action Recognition  & 10  \\ 
2019 & \cite{enrico2019dhp19}        & DHP19                   & 2,244   & 17     & EV       & Pose Estimation     & -   \\ 
2019 & \cite{wang2019ev}         & DVS128-Gait-Day         & 4,000   & 20     & EV       & Gait Recognition    & -   \\ 
2019 & \cite{wang2019ev}         & DVS128-Gait-Night       & 4,000   & 20     & EV       & Gait Recognition    & -   \\ 
2021 & \cite{liu2021event}        & DailyAction-DVS         & 1,440   & 15     & EV       & Action Recognition  & 12  \\ 
2022 & \cite{eddine2022gait3}       & Gait3                   & 168     & 56     & RGB - EV - TH & Gait Recognition & -   \\ 
2023 & \cite{gao2023action}         & THU-E-ACT-50            & 10,500  & 105    & EV & Action Recognition  & 50  \\ 
2023 & \cite{gao2023action}        & THU-E-ACT-50-CHL        & 2,330   & 18     & EV & Action Recognition  & 50  \\ 
2024 & \cite{gao2024hypergraph}        & THU-MV-E-ACT-50         & 31,500  & 105    & EV       & Action Recognition  & 50  \\ 
2025 & \cite{wang2025dailydvs}         & DailyDVS-200            & 22,000  & 46     & RGB - EV & Action Recognition  & 200 \\ \hline
\end{tabular}

\vspace{0.5cm} 

\begin{tabular}{ccccccc}
\multicolumn{7}{c}{\textbf{Face Datasets}} \\ \hline
\textbf{Year} & \textbf{Authors} & \textbf{Name}              & \textbf{\# Videos} & \textbf{\# People} & \textbf{Modalities} & \textbf{Application} \\ \hline
2016 & \cite{barua2016direct}        & -            & -       & 30     & EV          & Face Detection           \\ 
2019 & \cite{li2019lip}        & -            & 34,000       & 34     & EV-audio          & Lip Reading           \\ 
2020 & \cite{angelopoulos2021event} & -     & 24      & 24     & EV          & Eye gaze tracking \\ 
2020 & \cite{chen2020eddd}         & EDDD                  & 260     & 26     & EV          & Drowsiness    \\ 
2020 & \cite{lenz2020event}         & -             & 48      & 10     & EV          & Face Detection            \\ 
2020 & \cite{chen2020neurobiometric}                  & NeuroBiometric         & 180     & 45     & EV          & Authentication \\ 
2022 & \cite{banerjee2022gaze}     & -         & 3360    & 6      & RGB - EV      & Eye gaze tracking \\ 
2022 & \cite{becattini2022facial}   & -       & 455     & 25     & RGB - EV    & MER           \\ 
2022 & \cite{tan2022multigrained}         & DVS-Lip                  & 19,871     & 40     & EV          & Lip Reading    \\
2022 & \cite{moreira2022neuromorphic}      & NVSFD  & 436       & 40      & EV      & Identity Recognition             \\ 
2023 & \cite{bissarinova2023faces}  & FES                    & $\sim$4000 & 73    & EV          & Face Detection            \\ 
2023 & \cite{berlincioni2023neuromorphic}  & NEFER                  & 609     & 29     & RGB - EV    & MER           \\ 
2023 & \cite{kanamaru2023isolated}         & -                  & 1500     & 20     & EV          & Lip Reading    \\
2024 & \cite{adra2024beyond}                  & VETEX                  & 2506    & 30     & RGB - EV - TH & MER          \\ \hline
\end{tabular}
\label{tab:real_datasets}
\end{table}

\begin{table}[]
\footnotesize
\centering
\caption{Synthetic datasets for human-centered applications.}
\vspace{0.5cm}
\renewcommand{\arraystretch}{1.5}

\begin{tabular}{ccccccc}
\multicolumn{7}{c}{\textbf{Body Datasets}} \\ \hline
\textbf{Year} & \textbf{Authors} & \textbf{Name} & \textbf{\# Videos} & \textbf{\# People} & \textbf{Application} & \textbf{\# Classes} \\ \hline
2019 & \cite{wang2019ev}         & EV-CASIA-B         & 8,184  & 124   & Gait Recognition       & -   \\ 
2020 & \cite{bi2020graphtemporal}       & HMDB51-DVS         & 6,766  & -     & Action Recognition     & 51  \\ 
2020 & \cite{bi2020graphtemporal}      & UCF101-DVS         & 13,320 & -     & Action Recognition     & 101 \\ 
2022 & \cite{plizzari2022e2}        & N-EPIC-Kitchens    & 64     & -     & Action Recognition     & 8   \\ 
2023 & \cite{zou2023eventpose}          & SynEventHPD        & 9,197  & 47    & Pose Estimation        & -   \\ 
2023 & \cite{goyal2023moveenet}         & eH36m              & 748    & 7     & Pose Estimation        & -   \\ \hline
\end{tabular}

\vspace{0.5cm} 

\begin{tabular}{cccccc}
\multicolumn{6}{c}{\textbf{Face Datasets}} \\ \hline
\textbf{Year} & \textbf{Authors} & \textbf{Name} & \textbf{\# Videos} & \textbf{\# People} & \textbf{Application} \\ \hline
2022 & \cite{moreira2022neuromorphic}       & SynFED       & 6536     & 30    & Identity Recognition \\ 
2023 & \cite{barchid2023spikingfer}       & ADFES        & 198     & 22    & Face Expression Recognition \\ 
2023 & \cite{barchid2023spikingfer}      & Oulu-CASIA   & 480     & 80    & Face Expression Recognition \\ 
2023 & \cite{barchid2023spikingfer, Verschae2023EventGestureFacial}       & e-CK+  & 327     & 93    & Face Expression Recognition \\ 
2023 &\cite{barchid2023spikingfer, Verschae2023EventGestureFacial}       & e-MMI  & 2900+   & 75    & Face Expression Recognition \\ 
2023 & \cite{ryan2023real}          & -            & -       & 5     & Multitask Facial Analysis    \\ 
2024 & \cite{tan2024tackling}          & DVS-LRW100            & 107,664       & -     & Lip Reading \\ \hline
\end{tabular}

\label{tab:synthetic_datasets}
\end{table}

\begin{table}[]
\centering
\footnotesize
\caption{This table presents a summary of the works included in this survey that compare their event-based networks with RGB-trained models. Works are classified by their target application and the authors, year, and any reported improvement of event-based methods over RGB, if applicable are reported.}
\vspace{0.5cm}
\label{tab:rgb_vs_event}
\begin{tabular}{c|c|c|c|}
\cline{2-4}
\textbf{}                                                                                                                            & \textbf{Authors}   & \textbf{Findings}                                                                                                                                & \textbf{Improvements of Event}                                                                     \\ \hline
\multicolumn{1}{|c|}{\multirow{5}{*}{\textbf{\begin{tabular}[c]{@{}c@{}}Gait \\ recognition\end{tabular}}}}                          & \cite{wang2019ev}   & \begin{tabular}[c]{@{}c@{}}For viewing angles 72, 90 and 108, \\ EV-Gait performs better than RGB based approaches\end{tabular}                  & 3\% increase in accuracy                                                                           \\ \cline{2-4} 
\multicolumn{1}{|c|}{}                                                                                                               & \cite{sokolova2019human}    & \begin{tabular}[c]{@{}c@{}}Similar perfomances reported for \\ Event-based and RGB approaches\end{tabular}                                       & -                                                                                                  \\ \cline{2-4} 
\multicolumn{1}{|c|}{}                                                                                                               & \cite{wang2021event}   & \begin{tabular}[c]{@{}c@{}}For viewing angle 90 degrees,\\ EV-Gait-Graph performs better than RGB based approaches\end{tabular}                  & 0.5\% increase in accuracy                                                                         \\ \cline{2-4} 
\multicolumn{1}{|c|}{}                                                                                                               & \cite{eddine2022gait3}      & \begin{tabular}[c]{@{}c@{}}Advantage of event data over RGB and thermal for \\ gait recognition\end{tabular}                                     & 2\% increase in accuracy                                                                           \\ \cline{2-4} 
\multicolumn{1}{|c|}{}                                                                                                               & \cite{tao2024gaitspike}        & \begin{tabular}[c]{@{}c@{}}They report the advantage of event over RGB \\ across all different rotation angles for gait recognition\end{tabular} & Up to 14\% increase in accuracy                                                                    \\ \hline
\multicolumn{1}{|c|}{\multirow{2}{*}{\textbf{\begin{tabular}[c]{@{}c@{}}Action \\ recognition\end{tabular}}}}                        & \cite{plizzari2022e2}    & \begin{tabular}[c]{@{}c@{}}Event data can surpass RGB for action recognition\\  in unseen scenarios on test data\end{tabular}                    & 4\% increase in accuracy                                                                           \\ \cline{2-4} 
\multicolumn{1}{|c|}{}                                                                                                               & \cite{de2023eventtransact}  & \begin{tabular}[c]{@{}c@{}}Event surpass RGB action recognition models \\ in different setups\end{tabular}                                       & Up to 14\% increase in accuracy                                                                    \\ \hline
\multicolumn{1}{|c|}{\multirow{2}{*}{\textbf{\begin{tabular}[c]{@{}c@{}}Pose \\ estimation\end{tabular}}}}                           & \cite{goyal2023moveenet}       & Pose estimation from event data surpasses RGB data                                                                                               & Up to 5\% increase in accuracy                                                                     \\ \cline{2-4} 
\multicolumn{1}{|c|}{}                                                                                                               & \cite{kohyama20243d}     & \begin{tabular}[c]{@{}c@{}}Event does not suffer from motion blur as RGB does\\ for 3D-based pose estimation\end{tabular}                        & \begin{tabular}[c]{@{}c@{}}Error (in mm) is divided by 5 \\ in certain scenarios\end{tabular}      \\ \hline
\multicolumn{1}{|c|}{\multirow{2}{*}{\textbf{\begin{tabular}[c]{@{}c@{}}Face \\ detection\end{tabular}}}}                            & \cite{barua2016direct}       & Comparable results to Viola-Jones face detector                                                                                                  & -                                                                                                  \\ \cline{2-4} 
\multicolumn{1}{|c|}{}                                                                                                               & \cite{ryan2023real}   & \begin{tabular}[c]{@{}c@{}}Traditional RGB models perform better on RGB images\\  than on their simulated event data counterpart\end{tabular}    & -                                                                                                  \\ \hline
\multicolumn{1}{|c|}{\textbf{\begin{tabular}[c]{@{}c@{}}Lip \\ reading\end{tabular}}}                                                & \cite{kanamaru2023isolated}     & They combined event and RGB modalities for lip reading                                                                                           & -                                                                                                  \\ \hline
\multicolumn{1}{|c|}{\multirow{5}{*}{\textbf{\begin{tabular}[c]{@{}c@{}}Microexpression\\ and emotion \\ recognition\end{tabular}}}} & \cite{becattini2022facial}   & \begin{tabular}[c]{@{}c@{}}Event data overperforms RGB for detecting three\\ types of expressions: Positive, Neutral, Negative\end{tabular}      & Up to 9\% increase in accuracy                                                                     \\ \cline{2-4} 
\multicolumn{1}{|c|}{}                                                                                                               & \cite{berlincioni2023neuromorphic}  & \begin{tabular}[c]{@{}c@{}}Event overperforms RGB in the prediction of \\ seven different emotions\end{tabular}                                  & Up to 15\% increase in accuracy                                                                    \\ \cline{2-4} 
\multicolumn{1}{|c|}{}                                                                                                               & \cite{xiao2024estme}         & Event and RGB are merged as input to the network                                                                                                 & 1\% increase in accuracy                                                                           \\ \cline{2-4} 
\multicolumn{1}{|c|}{}                                                                                                               & \cite{cultrera2024spatio}   & \begin{tabular}[c]{@{}c@{}}For the estimation of some action units event data\\ delivers better performance\end{tabular}                         & \begin{tabular}[c]{@{}c@{}}For 6 out of 24 action units\\ event data is more accurate\end{tabular} \\ \cline{2-4} 
\multicolumn{1}{|c|}{}                                                                                                               & \cite{adra2024beyond}         & \begin{tabular}[c]{@{}c@{}}Event data gives more information than RGB \\ for microexpression recognition\end{tabular}                            & Up to 13\% increase in accuracy                                                                    \\ \hline
\end{tabular}
\end{table}


\end{document}